\documentclass[11pt]{article}

\usepackage[final]{acl}

\usepackage{times}
\usepackage{latexsym}
\usepackage{amsmath}
\usepackage{pifont}
\usepackage{subcaption}
\usepackage{multirow}
\usepackage{tabularx}
\usepackage{makecell}
\usepackage{booktabs}
\usepackage[T1]{fontenc}
\usepackage[utf8]{inputenc}

\usepackage{microtype}

\usepackage{graphicx}

\title{Beyond Single-Negative Preference: Multi-Negative DPO for LLM-Centric Historical Entity Linking}

\author{
  Tien Nam Nguyen$^1$ \quad
  Emanuela Boros$^1$ \quad
  Ahmed Hamdi$^2$ \quad \\
  {\bf Adam Jatowt}$^3$ \quad
  {\bf Mickaël Coustaty}$^1$ \quad
  {\bf Antoine Doucet}$^{1,4}$ \quad \\ 
  $^1$L3i, University of La Rochelle, La Rochelle, France \\
  $^2$University of Toulouse, IRIT, Toulouse, France \\
  $^3$University of Innsbruck, Austria \\
  $^4$FRI, University of Ljubljana, Ljubljana, Slovenia \\
}

\begin{document}
\maketitle
\begin{abstract}
% This document is a supplement to the general instructions for *ACL authors. It contains instructions for using the \LaTeX{} style files for ACL conferences.
% The document itself conforms to its own specifications, and is therefore an example of what your manuscript should look like.
% These instructions should be used both for papers submitted for review and for final versions of accepted papers.
Large language models (LLMs) have recently shown promise for historical entity linking, but preference optimization for this task is often formulated with only one negative candidate per training instance. This discards information from the remaining candidates retrieved for the same mention. We introduce multi-negative direct preference optimisation (MDPO), a reference-based pairwise objective that compares the correct entity with all valid rejected candidates associated with each mention. MDPO preserves the Bradley-Terry formulation of DPO while exploiting the complete candidate set through masked, length-normalised sequence scores. We evaluate MDPO on \texttt{hipe-2020} and \texttt{newseye}, covering French, German, English, Swedish, and Finnish historical newspaper text. Experiments show that MDPO improves over supervised fine-tuning and single-negative DPO, with particularly strong gains for NIL mentions, semantic ambiguity, OCR noise, and historically difficult names. Further analyses disentangle candidate-generation and selection errors, showing that candidate retrieval remains a key bottleneck for end-to-end entity linking. These results demonstrate that incorporating all within-instance negative candidates is a simple and effective improvement for LLM-based historical entity linking.

\end{abstract}

\section{Introduction}

Entity linking (EL) maps textual mentions to their corresponding entities in a knowledge base and is a core component of information extraction, question answering, digital library access, and knowledge base population. Standard EL systems usually decompose the task into candidate generation and entity disambiguation or ranking \citep{sevgili2022neural}: candidate generation retrieves plausible entities for a mention, while the ranking stage selects the most appropriate entity given the context. Recent neural approaches have improved both stages using dense retrieval, bi-encoders, cross-encoders, and generative models, including BLINK \citep{wu2019zero}, ELQ \citep{li2020efficient}, GENRE \citep{decao2021autoregressive,decao2022multilingual}, ReFinED \citep{ayoola2022refined}, and LLMAEL \citep{xin2025llmael}.

Despite these advances, EL remains difficult in historical and archival documents, which often contain OCR noise, spelling variations, obsolete or multilingual forms, incomplete metadata, temporally ambiguous references, and mentions that are absent from the target knowledge base. These factors make candidate retrieval, disambiguation, and NIL prediction especially challenging. The \texttt{hipe-2020} dataset, introduced through the HIPE shared tasks, highlights the difficulty of robust named entity recognition (NER) and linking (EL) in multilingual historical documents \citep{ehrmann2020introducing,ehrmann2020extended}. Complementarily, the \texttt{newseye} dataset provides multilingual historical newspaper material in French, German, Finnish, and Swedish. Together, these benchmarks show that historical EL requires methods that go beyond surface-form matching and account for noisy, incomplete, and historically situated contexts.

LLMs offer new opportunities for this setting because they combine broad parametric knowledge with contextual reasoning and generative capabilities. Recent work has explored LLMs for EL through prompting, instruction tuning, context augmentation, and selective reranking \citep{xiao2023instructed,ding2024entgpt,vollmers2025contextual,xin2025llmael,li2025leveraging}. In historical EL, LLMs have also been used to improve NIL prediction and candidate selection, often in combination with traditional retrievers \citep{santini2026confidence}. However, most existing approaches still treat LLMs as auxiliary components: they augment contexts, rerank externally retrieved candidates, or predict from a fixed candidate set. This limits their ability to generate plausible candidates and does not directly model the multi-candidate nature of entity disambiguation.

In this paper, we propose an LLM-centric framework for historical entity linking that uses LLMs in both candidate retrieval and entity selection. First, we introduce an LLM-guided candidate retrieval strategy, where an LLM generates plausible entity candidates from a mention and its surrounding context. The generated candidates are validated against a knowledge base and combined with alias-based retrieval, allowing the system to exploit both contextual generation and high-recall lexical lookup. This is particularly useful for historical documents, where OCR noise, spelling variation, and archaic surface forms can make conventional retrieval brittle.

Second, we reformulate entity selection as a preference-learning problem. Standard DPO trains a model from one chosen--rejected comparison \citep{rafailov2023direct}, whereas EL naturally supplies several hard competitors for the same mention. We introduce a multi-negative DPO objective that preserves the reference-based pairwise formulation while optimising all valid gold-versus-negative comparisons from each retrieved set.

Third, we evaluate the proposed framework on two multilingual historical-document benchmarks: \texttt{hipe-2020}, introduced for NER and EL in historical newspapers \citep{ehrmann2020introducing,ehrmann2020extended}, and \texttt{newseye}, a multilingual historical newspaper dataset covering French, German, Finnish, and Swedish material \citep{hamdi2021multilingual}. Our evaluation compares supervised fine-tuning, standard DPO, prompting-based baselines, and the proposed multi-negative DPO formulation, and provides ablation studies on retrieval, prompt design, context representation, and model choice.

\section{Related work}\label{sec:related-work}
% work in progress

\paragraph{LLM-based entity linking.}
EL is commonly formulated as a two-stage task: candidate generation followed by entity disambiguation. Traditional systems rely on lexical matching, prior probabilities, dense retrieval, and supervised ranking models, while recent work increasingly explores LLMs for candidate generation, context enrichment, and entity selection. INSGENEL \cite{xiao2023instructed} shows that instruction-tuned decoder-only models, combined with a retriever, can perform effective entity linking while reducing the cost of full sequence generation. EntGPT \cite{ding2024entgpt} demonstrates the potential of LLMs for entity disambiguation through prompting and instruction tuning, using LLMs to generate auxiliary information and select the correct entity from a candidate set. Hybrid methods such as ARTER \cite{li2025leveraging} combine fast entity linkers with targeted LLM reasoning, routing only difficult mentions to the LLM to balance accuracy and efficiency.

Another closely related direction uses LLMs to enrich the input context. \citet{vollmers2025contextual} propose contextual augmentation, where ambiguous mentions are expanded into more explicit Wikipedia-like titles using surrounding context. Similarly, LLMAEL \cite{xin2025llmael} uses LLMs as plug-and-play context augmenters and fuses the generated context with existing EL models such as BLINK \citep{wu2019zero,li2020efficient}, GENRE \cite{decao2021autoregressive,decao2022multilingual}, and ReFinED \cite{ayoola2022refined}. These approaches show that LLMs can help resolve ambiguity by adding missing contextual or world knowledge. Our method shares this intuition, but instead of only augmenting the mention context or reranking candidates, we prompt the LLM to generate plausible entity candidates, validate them against the knowledge base, augment them with alias-based retrieval, and then train an LLM-based selector using preference learning.

% \paragraph{Entity linking in multilingual, historical, and domain-specific settings.}
% Entity linking is especially difficult in multilingual, historical, and domain-specific corpora, where mentions may be noisy, rare, temporally ambiguous, or absent from standard knowledge bases. BELA \cite{plekhanov2023multilingual} addresses multilingual end-to-end entity linking with a large Wikipedia- and Wikidata-based entity index, jointly modelling mention detection, disambiguation, and rejection. In historical entity linking, \citet{santini2026confidence} propose MHEL-LLaMo, an unsupervised multilingual framework that combines a multilingual bi-encoder for candidate retrieval with an instruction-tuned LLM for NIL prediction and candidate selection. Their approach uses confidence scores to distinguish easy from hard cases and applies the LLM only to difficult mentions, reducing computational cost while improving robustness in historical documents. In historical and cultural heritage domains, \citet{graciotti2025musical} study musical heritage entity linking and highlight the importance of low-popularity entities, temporal constraints, NIL prediction, and OCR noise. \citet{cadavid2023evaluating} similarly show the limitations of generic entity linkers on museum collection data and demonstrate the value of domain-specific adaptation.

\paragraph{Entity linking in multilingual, historical, and domain-specific settings.}
Entity linking is especially difficult in multilingual, historical, and domain-specific corpora, where mentions may be noisy, rare, temporally ambiguous, or absent from standard knowledge bases. BELA \cite{plekhanov2023multilingual} addresses multilingual end-to-end entity linking with a large Wikipedia- and Wikidata-based index, jointly modeling mention detection, disambiguation, and rejection. For historical EL, \citet{santini2026confidence} propose MHEL-LLaMo, an unsupervised multilingual framework that combines bi-encoder retrieval with an instruction-tuned LLM for NIL prediction and candidate selection, applying the LLM mainly to difficult cases. In cultural heritage domains, studies on musical heritage and museum collections highlight the importance of low-popularity entities, temporal constraints, NIL prediction, OCR noise, and domain-specific adaptation \citep{graciotti2025musical,cadavid2023evaluating}.

\paragraph{Preference optimisation.}
DPO learns a reference-relative Bradley--Terry objective from one chosen--rejected pair \citep{rafailov2023direct}, whereas SimPO replaces the reference model with a length-normalised, reference-free reward \citep{meng2024simpo}. Multi-response methods address richer supervision: MPPO supports arbitrary negative samples and studies pointwise, pairwise, and listwise variants using average-likelihood rewards \citep{xie-etal-2025-mppo}, while LiPO couples candidates through a learning-to-rank loss over ranked response lists \citep{liu-etal-2025-lipo}. Our EL data instead provide one gold entity and an unordered set of retrieved competitors. We therefore retain DPO's reference-relative score, preserve the shared mention-level candidate structure, and aggregate independent gold-versus-negative terms. Unlike LiPO, our loss has no shared listwise normalisation; unlike single-negative DPO, it does not discard the remaining competitors. Our contribution is this task-specific construction and implementation rather than a new listwise objective. %We compare directly with single-negative DPO, while controlled comparisons with MPPO and LiPO remain future work.

\begin{figure*}[ht]
    \centering
    \includegraphics[width=.99\linewidth]{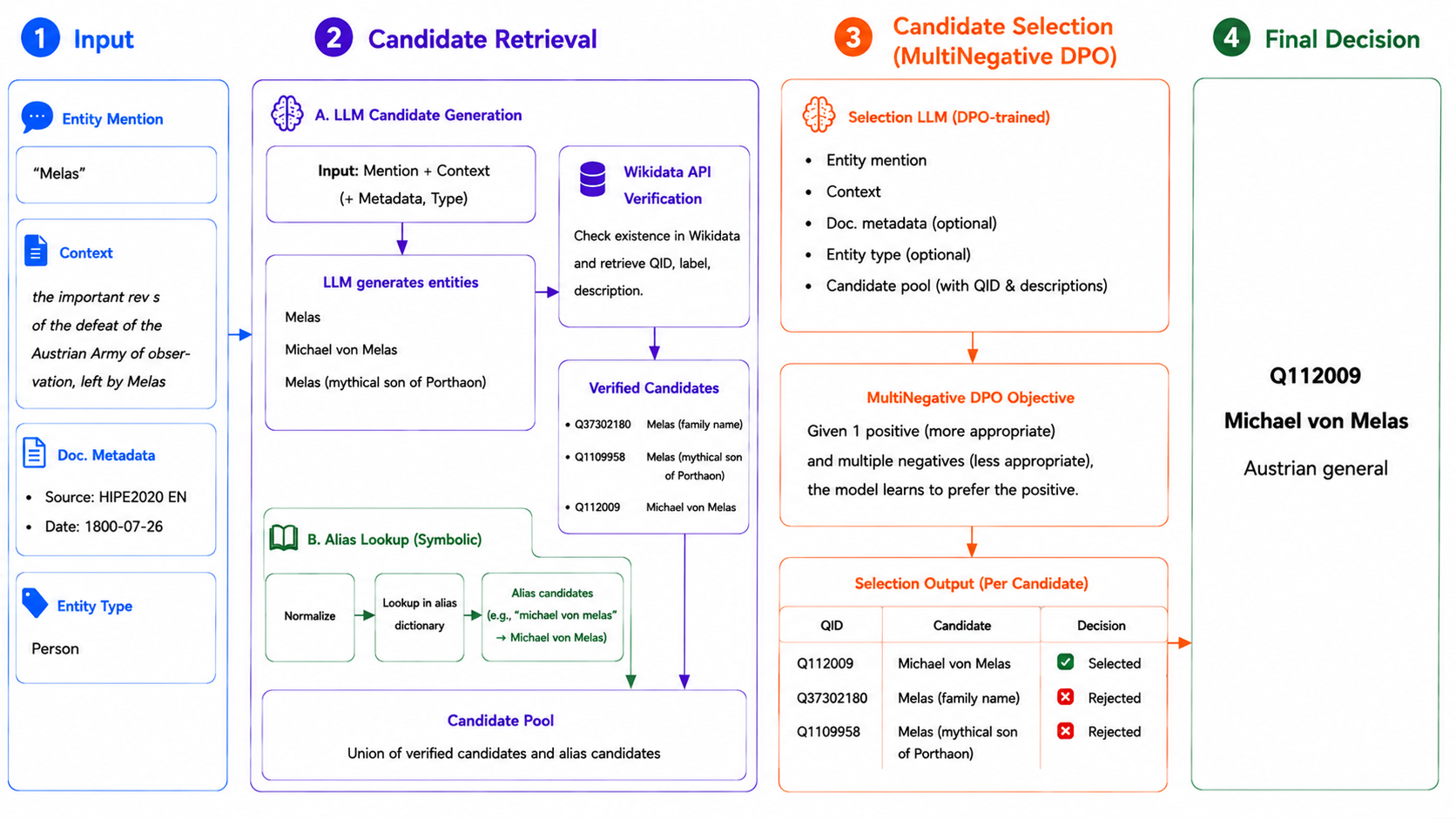}
    \caption{Overview of the proposed LLM-centric entity linking pipeline.}
    \label{fig:Overview-LLMs-NEL}
\end{figure*}

% Our work targets historical and archival documents, where OCR noise, spelling variation, incomplete knowledge bases, and NIL mentions make both candidate generation and entity selection particularly challenging. 
% Unlike previous approaches that mainly rely on contextual augmentation, dense retrieval, or constrained candidate filtering, we use LLMs as an integral part of both candidate generation and entity selection. We also propose a multi-negative preference-learning objective that better matches the structure of EL, where the correct entity must be selected from several competing candidates rather than compared against a single negative example.
% Unlike prior work that uses LLMs mainly as auxiliary components, our framework makes them central to retrieval and selection, using a multi-negative preference objective tailored to EL.

\section{Methodology}\label{sec:methodology}

Entity linking is naturally a multi-candidate decision problem: for each mention $m$ appearing in context $c$, the model must select the corresponding entity $e$ from a knowledge base $\mathcal{K}$ or abstain with a special \textsc{NIL} label when no valid entity exists. Formally, the task is defined as learning a function:
\begin{equation}
    f(m, c; \mathcal{K}) \rightarrow e,\quad e \in \mathcal{K} \cup \{\textsc{NIL}\}.
\end{equation}
This structure supplies multiple valid preference comparisons for each mention. We retain the gold entity and all hard negatives or NIL alternatives from the same retrieved set, rather than sampling only one negative. The training objective remains pairwise; ``multi-negative'' describes the within-instance data structure and aggregation, not a listwise normalisation.

We propose an LLM-centric entity linking pipeline that integrates candidate retrieval and entity selection through structured interaction with LLMs, as shown in Figure~\ref{fig:Overview-LLMs-NEL}. The pipeline consists of two main stages: (1) LLM-guided candidate retrieval and (2) LLM-based entity selection.

% \subsection{Task Definition}

% Entity Linking (EL) aims to map a textual mention $m$ appearing in context $c$ to a corresponding entity $e$ in a knowledge base $\mathcal{K}$. Formally, the task is defined as learning a function:
% \begin{equation}
%     f(m, c; \mathcal{K}) \rightarrow e,\quad e \in \mathcal{K}.
% \end{equation}
% For settings in which a mention may not correspond to any entity in the knowledge base, the output space can be extended with a special \textsc{NIL} label.

% Conventional EL systems typically decompose this problem into two stages: (1) candidate generation and (2) .Neural approaches often instantiate this pipeline using learned mention and entity representations, including bi-encoder retrieval followed by cross-encoder re-ranking \cite{ganea-hofmann-2017-deep,kolitsas-etal-2018-end,wu-etal-2020-scalable}. More recent work has also explored generative entity retrieval, where entity names are generated directly, as well as prompt-based LLM methods for ranking and few-shot entity linking \cite{decao2020autoregressive,qin-etal-2024-large,liu-etal-2024-onenet,ding2024entgpt}. 
% many existing approaches still rely heavily on an initial retrieval step or use LLMs mainly as external prompting or re-ranking modules. In this work, we propose an LLM-centric entity linking pipeline that exploits LLMs not only for selection, but also for candidate generation in noisy and historically variable text.

\subsection{LLM-guided Candidate Retrieval}
Traditional candidate retrieval methods for entity linking usually construct a small set of possible entities using lexical or heuristic signals, such as alias dictionaries, mention entity priors, inverted indices, and approximate string matching \cite{bunescu-pasca-2006-using,cucerzan-2007-large,ratinov-etal-2011-local,robertson2009probabilistic,zhang-etal-2022-knowledge}. While efficient, these methods are sensitive to surface-form mismatch. This is particularly problematic in historical corpora, where mentions may contain OCR errors, spelling variants, archaic forms, abbreviations, or semantic shifts \cite{provatorova2020named,hamdi2023impact}. If the correct entity is not retrieved at this stage, the subsequent selection model might not find it.

To improve recall, we adopt an LLM-guided candidate retrieval strategy. Instead of relying only on lexical similarity, we prompt the LLM with the mention $m$ and its surrounding context $c$ to generate semantically plausible candidate strings, including canonical names, aliases, normalised spellings, and historically plausible variants. These generated strings are not treated as final predictions; rather, they are used to expand retrieval over the knowledge base.

Formally, we define the LLM-generated query set as:
\begin{equation}
    \mathcal{C}_{\mathrm{LLM}}(m,c) =
    \mathrm{Retrieve}_{\mathcal{K}}\!\left(\mathrm{LLM}_{\mathrm{gen}}(m,c)\right),
\end{equation}
where $\mathcal{C}_{\mathrm{LLM}}(m,c)$ denotes the knowledge-base candidates retrieved from the generated strings.
Because LLMs may generate hallucinated or non-existent entity names, we validate the retrieved candidates against the knowledge base:
\begin{equation}
\begin{split}
\mathcal{C}_{\mathrm{valid}}(m,c)
= \{\, e \in \mathcal{C}_{\mathrm{LLM}}(m,c) \mid
\operatorname{Valid}_{\mathcal{K}}(e) \,\}.
\end{split}
\end{equation}
Here, $\operatorname{Valid}_{\mathcal{K}}$ checks whether the candidate corresponds to a valid entity identifier in the knowledge base.

To further improve recall, we build an alias dictionary independently for each benchmark using only its training split. Each normalised mention surface form is mapped to all of its annotated QID(s); one-to-many mappings are retained, and no external or manually curated aliases are used. Duplicate alias and LLM candidates are removed before the two sources are merged. Thus, test-only mention--entity associations cannot enter the lookup table. We define:
\begin{equation}
    \mathcal{C}_{\mathrm{alias}}(m) =
    \mathrm{AliasLookup}_{\mathcal{K}}(m).
\end{equation}
The final candidate set combines validated LLM-guided candidates with alias-based matches:
\begin{equation}
    \mathcal{C}(m,c) =
    \mathcal{C}_{\mathrm{valid}}(m,c)
    \cup
    \mathcal{C}_{\mathrm{alias}}(m).
\end{equation}

This approach uses the LLM to expand the search space while relying on the knowledge base to ensure candidate validity. Thus, the system improves recall without directly accepting unsupported LLM-generated entities. %Table~\ref{tab:retrieval_recall_prompt} shows alias gains of $+1.5/+1.5$ points on hipe-2020 FR/DE and $+1.2/+2.9$ points on NewsEye FR/DE.

\subsection{LLM-based Entity Selection}

% \begin{figure*}
%     \centering
%     \includegraphics[width=0.9\linewidth]{LLMs-Retrieval.png}
%     \caption{LLMs-guided Candidate Retrieval with alias}
%     \label{fig:LLMs-guied Retrieval}
% \end{figure*}
In the second stage, the LLM performs joint reasoning over the candidate set to select the most appropriate entity. Given a mention $m$, its context $c$, and a candidate set $\mathcal{C}$ with associated descriptions, the model predicts the most likely entity:
\begin{equation}
\hat{e} = \arg\max_{e \in \mathcal{C}} P(e \mid m, c, \mathcal{C})
\end{equation}

% combined this with advantages
% \paragraph{Limitations of Standard DPO.}
% A natural way to train LLMs for this task is to formulate entity disambiguation as a preference learning problem, where the correct entity $e^{+}$ should be preferred over incorrect candidates. DPO provides a principled framework for such training. However, standard DPO operates on pairwise comparisons involving only one positive and one negative sample at a time.  This formulation is insufficient for EL, where the model must discriminate among a set of many competing candidates. Relying on a single negative per update fails to capture the relative structure of the candidate set and leads to weak supervision, especially when multiple hard negatives exist.

% \paragraph{Formulating Entity Linking as Preference Learning.}
\paragraph{EL as preference learning.}
We reformulate entity linking as a preference learning problem over the candidate set $\mathcal{C}$. Given a gold entity $e^{+}$ and a set of negative candidates $\{e_1^{-}, e_2^{-}, \dots, e_n^{-}\}$, the objective is to learn preferences such that: $e^{+} \succ e_i^{-}, \quad \forall i \in \{1, \dots, n\}$.

% \begin{equation}
% e^{+} \succ e_i^{-}, \quad \forall i \in \{1, \dots, n\}
% \end{equation}

\paragraph{Multi-negative DPO.}
Let $x_b=(m_b,c_b,\mathcal{C}_b)$ and $\bar{\ell}_{\pi}(e\mid x)=|e|^{-1}\log\pi(e\mid x)$ be the length-normalised sequence log-likelihood. For instance $b$ with $n_b$ valid negatives, we optimise:
\begin{equation}
\label{eq:mdpo}
\begin{aligned}
r_\theta(e\mid x)
&=
\bar{\ell}_{\pi_\theta}(e\mid x)
-
\bar{\ell}_{\pi_{\mathrm{ref}}}(e\mid x), \\[2pt]
\Delta_{b,i}
&=
r_\theta(e_b^+\mid x_b)
-
r_\theta(e_{b,i}^-\mid x_b), \\[2pt]
\mathcal{L}_{\mathrm{MDPO}}
&=
-\frac{1}{\sum_{b=1}^{B} n_b}
\sum_{b=1}^{B}
\sum_{i=1}^{N}
M_{b,i}
\log \sigma\!\left(\beta\Delta_{b,i}\right), \\[2pt]
n_b
&=
\sum_{i=1}^{N} M_{b,i}.
\end{aligned}
\tag{7}
\end{equation}
% \begin{equation}
% \mathcal{L}_{\text{MDPO}} = - \sum_{i=1}^{n} \log \sigma \left( \beta \left( 
% \log P(e^{+} \mid m, c) - \log P(e_i^{-} \mid m, c) \right) \right)
% \end{equation}
Here $\pi_\theta$ is the LoRA-adapted policy, $\pi_{\mathrm{ref}}$ is the same pretrained backbone with its adapters disabled, and $\beta>0$ controls preference strength. The batch and tensor notation makes the implementation explicit. $B$ is the number of mentions in a minibatch, $N$ is the maximum number of valid rejected candidates in that minibatch, and $L$ is the maximum token length used when scoring one entity response. The chosen responses therefore have shape $[B,L]$, while the rejected responses are padded to $[B,N,L]$. $M_{b,i}\in\{0,1\}$ is a validity mask: it is one when the $i$-th negative belongs to mention $b$ and zero for padding or an unavailable comparison. Hence $n_b$ counts only real negatives, and the denominator averages over valid comparisons rather than padded tensor positions.

For each mention, the trainer computes the length-normalised chosen score once and reuses it for all valid negatives. It then computes one independent margin $\Delta_{b,i}$ for each gold--negative pair and applies the binary Bradley-Terry likelihood to that pair. A positive margin means that the policy improves the gold entity's reference-relative score more than the corresponding negative's. Normalisation by $\sum_b n_b$ gives every valid comparison equal weight, while an instance with more hard negatives contributes more pairwise evidence. Crucially, $N$ and the $[B,N,L]$ tensor are implementation dimensions, not a listwise probability distribution: there is no softmax over candidates, no negative-to-negative interaction, and no shared candidate-set normalisation. Thus, ``multi-negative'' means that every retrieved competitor supplies supervision for the same mention, whereas the objective remains a sum of ordinary reference-based pairwise DPO terms. If $n_b=1$, Eq.~\ref{eq:mdpo} becomes the single-negative version of this length-normalised, reference-based DPO loss. %It is identical to canonical DPO only when the standard-DPO baseline uses the same length normalization.

\section{Preference Data Construction}\label{sec:preference-data}
To train our model with the multi-negative DPO objective, we construct preference data automatically from standard entity linking annotations. Each instance consists of a mention $m$, context $c$, and a ground-truth entity $e^{+}$, which can be either a valid knowledge base entity (non-NIL) or NIL.

We first apply our LLM-based candidate generation pipeline to obtain a candidate set $\mathcal{C}$. Based on the relationship between $e^{+}$ and $\mathcal{C}$, we consider four cases:

\paragraph{Case 1: Non-NIL with Correct Candidate Present.}
If $e^{+} \neq \text{NIL}$ and $e^{+} \in \mathcal{C}$, we construct preference pairs where the gold entity is preferred over all other candidates:
% \begin{equation}
% (e^{+}, e^{-}), \quad \forall e^{-} \in \mathcal{C} \setminus \{e^{+}\}
% \end{equation}
\[
(e^+, e^-), \qquad \forall e^- \in C \setminus \{e^+\}
\]
\paragraph{Case 2: Non-NIL with Missing Candidate.}
If $e^{+} \neq \text{NIL}$ but $e^{+} \notin \mathcal{C}$, we explicitly add $e^{+}$ into the candidate set:
% \begin{equation}
% \mathcal{C} \leftarrow \mathcal{C} \cup \{e^{+}\}
% \end{equation}
\[
C \leftarrow C \cup \{e^+\}
\]
and construct preference pairs as in Case 1. This ensures the model always observes the correct entity during training.

\paragraph{Case 3: NIL with Non-empty Candidate Set.}
If $e^{+} = \text{NIL}$ and $\mathcal{C} \neq \emptyset$, we treat NIL as the correct choice and prefer it over all retrieved candidates:
% \begin{equation}
% (\text{NIL}, e^{-}), \quad \forall e^{-} \in \mathcal{C}
% \end{equation}
\[
(\mathrm{NIL}, e^-), \qquad \forall e^- \in C
\]
\paragraph{Case 4: NIL with NIL Candidate Present.}
If NIL is already included in $\mathcal{C}$, we retain it as the preferred candidate and construct preference pairs against all non-NIL entities.

\paragraph{DPO Formatting.}
Each record stores one prompt $(m,c,\mathcal{C})$, one \textit{chosen} entity, and the complete list of valid \textit{rejected} entities. These comparisons remain grouped through collation and loss computation (Appendix~\ref{app:training-details}). Records with no valid rejected entity, including empty-candidate gold-NIL cases, provide no preference comparison and are excluded from DPO training.

% \paragraph{Discussion.}

\section{Experiments}\label{sec:experiments}

% \subsection{Experimental Setup}

% We evaluate our proposed LLM-based entity linking pipeline on benchmark datasets for news and historical entity linking. Our setup follows a standard two-stage framework, where candidate generation is performed using a large language model, and entity selection is trained using the proposed multi-negative DPO objective.

% We consider both entity disambiguation and NIL prediction settings, reflecting realistic scenarios where mentions may not correspond to any entity in the knowledge base.

%\subsection{Datasets}

%We conduct experiments on the following datasets:

%\paragraph{NewsEL.}
%A news-domain entity linking dataset containing mentions annotated with corresponding entities in a knowledge base.

%\paragraph{HIPE 2020.}
%The HIPE 2020 dataset focuses on historical documents and presents additional challenges such as ambiguous mentions and incomplete knowledge base coverage.

%These datasets allow us to evaluate the robustness of our approach across both contemporary and historical domains, as well as its ability to handle NIL cases.

\subsection{Datasets}

We conduct experiments on the following datasets:
\paragraph{\texttt{hipe-2020}} \cite{ehrmann2020introducing} dataset provides a benchmark for NER and EL on multilingual historical newspapers. The dataset contains $17,553$ named entity mentions distributed across German, French, and English historical newspaper corpora, with annotations aligned to Wikidata for the entity linking task. The corpus includes $8,205$ location entities, $6,113$ person entities, $1,971$ organisation entities, $679$ temporal expressions, and $585$ product entities, reflecting a diverse range of semantic categories in historical documents.

\paragraph{\texttt{newseye}} \cite{hamdi2021multilingual} dataset provides a suitable benchmark for the entity linking task, as it includes $30,580$ named entities, among which $6,704$ are aligned with Wikidata. This explicit grounding of entities in a knowledge base enables the evaluation of linking approaches in a realistic, multilingual setting. The datasets consider four languages (French, German, Finnish, and Swedish), covering a diverse range of linguistic characteristics. The corresponding subsets contain $12,473$ entities for French, $12,818$ for German, $2,572$ for Finnish, and $2,717$ for Swedish, mainly spanning persons, locations, and organisations. The \texttt{newseye} dataset enables robust cross-lingual comparisons with the \texttt{HIPE} dataset, which is also used in our study. Both of them ensure consistency in annotation schemes.

\begin{table*}[ht]
\centering
\small
\renewcommand{\arraystretch}{0.95}
\begin{tabular}{llccccccc}
\toprule
\textbf{Approach} & \textbf{Setup} 
& \multicolumn{3}{c}{\texttt{hipe-2020}} 
& \multicolumn{4}{c}{\texttt{newseye}} \\
\cline{3-9}
&  
& \textbf{FR} & \textbf{DE} & \textbf{EN} 
& \textbf{FR} & \textbf{DE} & \textbf{SV} & \textbf{FI} \\
\toprule
SBB~\citep{labusch2020named}  
& Wiki emb. + BERT/RF 
& 59.6 & 50.6 & 39.3 
& 44.4 & 43.1 & -- & -- \\

L3i~\citep{boros_robust_2020}
& Neural EL + postproc. 
& 60.2 & 48.1 & 54.6 
& -- & -- & -- & -- \\

MELHISSA~\citep{linhares2022melhissa}  
& Multiling. EL + filters 
& 63.0 & 57.3 & 59.7 
& 54.2 & 54.7 & 59.9 & \textbf{65.2} \\

BELA~\citep{plekhanov2023multilingual}  
& Bi-enc. + score 
& 58.0 & 55.0 & 41.8 
& 42.2 & 30.0 & 41.0 & 30.0 \\

MHEL-LLaMo~\citep{santini2026confidence}  
& BELA + LLM 
& 69.2 & 62.0 & 72.3 
& 66.2 & 55.6 & 52.1 & 50.9 \\

\midrule
Prompting & GPT-120B / GPT-20B 
& 65.7 & 59.6 &  65.2
&57.9  & 49.2  & 53.3 &60.4 \\

SFT 
& GPT-120B / Qwen3-14B 
& 60.3 & 55.0 & 46.2 
& 44.8 & 36.6 & 48.2 & 38.4 \\

DPO 
& GPT-120B / GPT-20B 
& 70.2 & 62.5 & 70.9 
& 63.5 & 52.8 & 62.8 & 63.5 \\

\textbf{Multi-DPO} 
& GPT-120B / GPT-20B 
& \textbf{72.4} & \textbf{65.6} & \textbf{73.8} 
& \textbf{68.7} & \textbf{59.2} & \textbf{65.0} & 62.7 \\
\bottomrule
\end{tabular}
\caption{Micro-F1 on \texttt{hipe-2020} and \texttt{newseye}.}
\label{tab:sota_hipe_newseye}
\end{table*}

% \subsection{Implementation Details}
\subsection{Models and Implementation}
We employ multiple LLMs across different components of the pipeline to balance retrieval quality and computational efficiency.

For candidate generation, we use a diverse set of open-source and large-scale LLMs, including GPT-20B, 120B \cite{agarwal2025gpt}, Qwen3.5-9B and 27B \cite{qwen35blog}, Qwen3-A30B \cite{yang2025qwen3}, Gemma4-31B \cite{gemma4_2026}. These models are used in a zero-shot setting to generate semantically relevant and diverse candidate entity names given a mention and its context. 

%Leveraging multiple models allows us to capture complementary knowledge and improve recall, particularly in noisy and multilingual settings.

For entity selection, we finetune Qwen3-14B \cite{yang2025qwen3}, GPT-20B \cite{agarwal2025gpt}, or Qwen3-32B \cite{yang2025qwen3} using the proposed multi-negative DPO. Details of the training and inference settings can be found in Appendix~\ref{app:training-details}.

This design choice decouples candidate generation from ranking, enabling the use of large models for high-recall retrieval while maintaining efficiency during downstream selection.

We implement structured prompting and output formatting using the OpenAI Agents framework\footnote{\url{https://github.com/openai/openai-agents-python}}, which supports schema-constrained generation and function calling for candidate generation, validation, and entity selection. Since the framework is not fully compatible with some fine-tuned Qwen-based models, we use direct prompting with manually enforced structured outputs while preserving the same interaction protocol and output schema. This ensures consistent structured outputs and reduces parsing errors during inference.

% \subsection{Baselines}
% \subsection{Compared Methods}
\subsection{Compared Systems} % and Training Objectives}

% \begin{itemize}
%     \item \textbf{Historical EL systems.} We report published results for SBB, L3i, MELHISSA, BELA, and MHEL-LLaMo. These systems cover lexical, neural, multilingual, and LLM-assisted approaches to entity linking in historical documents.

%     \item \textbf{LLM prompting.} We include a prompting-based setting in which an LLM selects an entity directly from the retrieved candidate list without fine-tuning.

%     \item \textbf{Supervised fine-tuning (SFT).} We train the same base model with standard supervised fine-tuning, optimizing it to generate the correct entity given the input $(m, c, \mathcal{C})$.

%     \item \textbf{DPO:} 
%     \begin{itemize}
%         \item \textbf{Standard}: We include a pairwise DPO variant with one positive and one negative candidate per training instance.
%         \item \textbf{Multi-negative}: Our proposed method extends DPO by comparing the gold entity against multiple negative candidates within each instance, providing a more structured training signal for entity selection.
%     \end{itemize}
    
% \end{itemize}
We compare our approach against both prior historical entity linking systems and several training configurations built on the same retrieval-augmented selection framework.

\paragraph{Prior historical EL systems.} We report published results for SBB, L3i, MELHISSA, BELA, and MHEL-LLaMo, which represent lexical, neural, multilingual, and LLM-assisted approaches for historical entity linking.

\paragraph{Our training configurations.}
All variants use the same retrieval-augmented setup and differ only in the selection strategy:

\begin{itemize}
   \item{LLM prompting.} The LLM selects an entity directly from the retrieved candidate list without task-specific fine-tuning.
    \item{Supervised fine-tuning (SFT).} The base model is fine-tuned to generate the correct entity given the input $(m, c, \mathcal{C})$.
    \item{DPO (standard).} A pairwise DPO setting using one positive and one negative candidate for each training instance.
    \item{DPO (multi-negative).} Our proposed approach extends DPO by contrasting the gold entity against multiple negative candidates within the same instance, providing a richer supervision signal for entity selection.

\end{itemize}
% We compare our method against prior historical entity linking systems, LLM prompting, supervised fine-tuning, and preference-optimization variants.

% \paragraph{Historical EL systems.}
% We report published results for representative historical and multilingual entity linking systems, including SBB, L3i, MELHISSA, BELA, and MHEL-LLaMo. These systems cover lexical, neural, multilingual, and LLM-assisted approaches to entity linking in historical documents.

% \paragraph{LLM prompting.}
% We include a prompting-based setting in which an LLM performs entity selection directly from the retrieved candidate list without fine-tuning.

% \paragraph{Supervised Fine-Tuning (SFT).}
% We train the same base model using standard supervised fine-tuning, where the model is optimized to generate the correct entity given the input $(m, c, \mathcal{C})$.

% \paragraph{Standard DPO.}
% We include a standard DPO variant that uses pairwise preference learning with one positive and one negative candidate per training instance.

% \paragraph{Multi-negative DPO.}
% Our proposed method extends DPO by incorporating multiple negative candidates per instance, providing a stronger and more structured training signal for entity selection.

\subsection{Evaluation Metrics}

Following the standard evaluation protocol of the HIPE 2020 shared task\footnote{\url{https://hipe-eval.github.io/HIPE-2022/evaluation}}, we use micro F1 as the primary evaluation metric for a fair comparison with previous work. In addition, since our framework contains an explicit candidate retrieval stage, we also report entity-level retrieval recall to evaluate the effectiveness of the retrieval component independently of the final entity selection stage.

% Micro F1 is computed over all mentions and measures the overall entity linking performance, including both entity disambiguation and NIL prediction.
% This metric is particularly suitable for our setting as it captures performance across imbalanced entity distributions and accounts for both precision and recall in a unified manner.

\section{Results}\label{sec:results}

\paragraph{Main Results.}

Table~\ref{tab:sota_hipe_newseye} reports micro-F1 scores on \texttt{hipe-2020} and \texttt{newseye}. We compare against prior historical EL systems reported in the literature, including SBB, L3i, MELHISSA, BELA, and MHEL-LLaMo, as well as our SFT, DPO, and Multi-DPO variants.

The results show that DPO substantially improves historical EL performance across both datasets. Standard DPO already outperforms SFT and original with prompting in most settings, suggesting that preference-based supervision is better suited than token-level generation loss or original knowledge of LLMs for retrieval-augmented entity selection. This is especially visible on \texttt{hipe-2020}-FR, \texttt{hipe-2020}-EN, and the \texttt{newseye} subsets.

Multi-DPO further improves upon standard DPO in most languages, achieving the best results on all \texttt{hipe-2020} subsets and on three of the four \texttt{newseye} subsets. The only exception is \texttt{newseye}-FI, where MELHISSA remains the strongest. These results suggest that modeling multiple preference relations provides a stronger ranking signal and improves robustness when the candidate set contains several plausible but incorrect entities.

% \begin{table}[h]
% \centering
% \scriptsize
% \setlength{\tabcolsep}{2.5pt}
% \begin{tabular}{lrrrr}
% \toprule
% \textbf{Error type} & $N$ & \textbf{Base} & \textbf{Finetuned} & \textbf{Gain (pp)} \\
% \midrule
% NIL / KB & 3636 & 48.4 & 62.1 & +13.7 \\
% Semantic & 801 & 59.4 & 67.5 & +8.1 \\
% OCR noise & 79 & 53.2 & 60.8 & +7.6 \\
% Other & 2993 & 68.2 & 71.8 & +3.6 \\
% Spelling / orthog. & 1194 & 63.7 & 64.9 & +1.2 \\
% Historical naming & 717 & 56.5 & 57.5 & +1.0 \\
% \bottomrule
% \end{tabular}
% \caption{Accuracy by heuristic error type across the seven evaluation subsets. ``Other'' contains cases not matched by the surface-form heuristics.}
%\label{tab:error_type_gain}
%\end{table}

\subsection{Oracle Entity Selection Analysis}
\begin{table}[ht]
\centering
\small
\setlength{\tabcolsep}{3pt}
\renewcommand{\arraystretch}{0.95}
\begin{tabular}{llp{1cm}p{.6cm}p{.5cm}rr}
\toprule
\textbf{Dataset} & \textbf{Lang.} & \textbf{Gold-in-set} & \textbf{Sel. acc.} & \textbf{Full} & \textbf{Oracle} & \textbf{Gain} \\
\midrule
\multirow{3}{*}{\texttt{hipe-2020}}
  & FR & 87.2 & 67.9 & 61.2 & 82.8 & +21.6 \\
  & DE & 83.4 & 64.9 & 56.0 & 79.3 & +23.3 \\
  & EN & 82.6 & 84.1 & 74.0 & 81.7 &  +7.7 \\
\midrule
\multirow{4}{*}{\texttt{newseye}}
  & FR & 88.5 & 67.1 & 55.7 & 71.9 & +16.2 \\
  & DE & 84.7 & 56.8 & 46.8 & 66.2 & +19.4 \\
  & FI & 88.2 & 53.0 & 63.1 & 85.1 & +22.0 \\
  & SV & 84.2 & 69.8 & 65.0 & 81.5 & +16.5 \\
\bottomrule
\end{tabular}
\caption{Conditional oracle analysis (\%). \textbf{Gold-in-set}: non-NIL gold retrieval recall; \textbf{Sel. acc.}: accuracy conditioned on gold retrieval; \textbf{Full}/\textbf{Oracle}: entity-level accuracy.}
\label{tab:oracle_analysis}
\end{table}

We isolate the remaining selection headroom using the candidate sets from the full retrieval pipeline. For each non-NIL mention whose gold QID is retrieved, the oracle replaces the model prediction with that QID; all other predictions, including NIL mentions, remain unchanged. Table~\ref{tab:oracle_analysis} reports gold-in-set recall among non-NIL mentions, selection accuracy conditional on successful retrieval, and full entity-level accuracy before and after the replacement.

% \begin{table}[t]
% \centering
% \scriptsize
% \setlength{\tabcolsep}{5pt}
% \begin{tabular}{lrrrrr}
% \toprule
% \textbf{Dataset} & \textbf{Gold-in-set} & \textbf{Sel. acc.} & \textbf{Full} & \textbf{Oracle} & \textbf{Gain} \\
% \midrule
% \texttt{hipe-2020}-FR & 87.2 & 67.9 & 61.2 & 82.8 & +21.6 \\
% \texttt{hipe-2020}-DE & 83.4 & 64.9 & 56.0 & 79.3 & +23.3 \\
% \texttt{hipe-2020}-EN & 82.6 & 84.1 & 74.0 & 81.7 &  +7.7 \\
% \texttt{newseye}-FR   & 88.5 & 67.1 & 55.7 & 71.9 & +16.2 \\
% \texttt{newseye}-DE   & 84.7 & 56.8 & 46.8 & 66.2 & +19.4 \\
% \texttt{newseye}-FI   & 88.2 & 53.0 & 63.1 & 85.1 & +22.0 \\
% \texttt{newseye}-SV   & 84.2 & 69.8 & 65.0 & 81.5 & +16.5 \\
% \bottomrule
% \end{tabular}
% \caption{Conditional oracle analysis (\%). \textbf{Gold-in-set}: non-NIL gold retrieval recall; \textbf{Sel. acc.}: accuracy conditioned on gold retrieval; \textbf{Full}/\textbf{Oracle}: entity-level accuracy.}
% \label{tab:oracle_analysis}
% \end{table}

Oracle gains range from 7.7 to 23.3 points. Thus, even with high non-NIL candidate recall, entity selection remains a substantial source of end-to-end error. The \texttt{newseye}-FI oracle uses all mentions, whereas conditional selection accuracy uses only retrieved non-NIL mentions; its high NIL fraction explains the apparent gap.

\subsection{Retrieval Analysis}

% \begin{table*}[ht]
% \centering
% \small
% \setlength{\tabcolsep}{4pt}
% \renewcommand{\arraystretch}{0.95}
% \begin{tabular}{lccccccc}
% \hline
% \multirow{2}{*}{\textbf{Candidate generation model}} 
% & \multicolumn{3}{c}{\texttt{hipe-2020}} 
% & \multicolumn{4}{c}{\texttt{newseye}} \\
% \cline{2-8}
% & \textbf{FR} & \textbf{DE} & \textbf{EN} 
% & \textbf{FR} & \textbf{DE} & \textbf{SV} & \textbf{FI} \\
% \hline
% Qwen3.5-9B~\citep{qwen35blog} 
% & 68.8 & 63.5 & 58.8 & 55.0 & 46.2 & 55.7 & 58.6 \\

% GPT-20B~\citep{agarwal2025gpt} 
% & 66.2 & 63.3 & 62.4 & 56.3 & 49.8 & 62.1 & 65.8 \\

% Qwen3.5-27B~\citep{qwen35blog} 
% & 75.0 & 72.9 & 61.5 & 55.8 & 50.1 & 62.8 & 59.3 \\

% Qwen3-A30B~\citep{yang2025qwen3} 
% & 72.2 & 66.7 & 60.9 & \textbf{59.2} & \textbf{51.7} & 11.4 & 24.4 \\

% Gemma4-31B~\citep{gemma4_2026} 
% & \textbf{76.1} & \textbf{73.6} & 64.9 & 55.7 & 50.8 & 59.4 & 63.1 \\

% GPT-120B~\citep{agarwal2025gpt} 
% & 73.1 & 65.8 & \textbf{66.2} & 58.9 & 51.0 & \textbf{64.1} & \textbf{69.6} \\
% \hline
% \end{tabular}
% \caption{Retrieval recall across datasets using different LLMs for candidate generation. Recall is computed as the proportion of mentions for which the gold entity appears in the retrieved candidate set.}
% \label{tab:retrieval_recall_base_llms}
% \end{table*}

%\paragraph{Baseline LLMs}

\paragraph{Backbone LLMs.}
Tables~\ref{tab:retrieval_recall_hipe} and~\ref{tab:retrieval_recall_newseye} compare candidate retrieval recall across backbone LLMs without alias augmentation, isolating each model's intrinsic candidate-generation ability. All models are evaluated using the same prompting strategy (Appendix~\ref{sec:appendix}).

\begin{table}[ht]
\centering
\small
\setlength{\tabcolsep}{6pt}
\renewcommand{\arraystretch}{0.95}
\begin{tabular}{lccc}
\toprule
\textbf{Candidate model} 
& \multicolumn{3}{c}{\texttt{hipe-2020}} \\
\cline{2-4}
& \textbf{FR} & \textbf{DE} & \textbf{EN} \\
\midrule
Qwen3.5-9B~\citep{qwen35blog} 
& 68.8 & 63.5 & 58.8 \\
GPT-20B~\citep{agarwal2025gpt} 
& 66.2 & 63.3 & 62.4 \\
Qwen3.5-27B~\citep{qwen35blog} 
& 75.0 & 72.9 & 61.5 \\
Qwen3-A30B~\citep{yang2025qwen3} 
& 72.2 & 66.7 & 60.9 \\
Gemma4-31B~\citep{gemma4_2026} 
& \textbf{76.1} & \textbf{73.6} & 64.9 \\
GPT-120B~\citep{agarwal2025gpt} 
& 73.1 & 65.8 & \textbf{66.2} \\
\bottomrule
\end{tabular}
\caption{Retrieval recall on \texttt{hipe-2020} using different LLMs for candidate generation.}
\label{tab:retrieval_recall_hipe}
\end{table}

\begin{table}[ht]
\centering
\small
\setlength{\tabcolsep}{6pt}
\renewcommand{\arraystretch}{0.95}
\begin{tabular}{lcccc}
\toprule
\textbf{Candidate model} 
& \multicolumn{4}{c}{\texttt{newseye}} \\
\cline{2-5}
& \textbf{FR} & \textbf{DE} & \textbf{SV} & \textbf{FI} \\
\midrule
Qwen3.5-9B
& 55.0 & 46.2 & 55.7 & 58.6 \\
GPT-20B
& 56.3 & 49.8 & 62.1 & 65.8 \\
Qwen3.5-27B
& 55.8 & 50.1 & 62.8 & 59.3 \\
Qwen3-A30B 
& \textbf{59.2} & \textbf{51.7} & 11.4 & 24.4 \\
Gemma4-31B
& 55.7 & 50.8 & 59.4 & 63.1 \\
GPT-120B
& 58.9 & 51.0 & \textbf{64.1} & \textbf{69.6} \\
\bottomrule
\end{tabular}
\caption{Retrieval recall on \texttt{newseye} using different LLMs for candidate generation.}
\label{tab:retrieval_recall_newseye}
\end{table}

The results show that model scale alone does not determine retrieval quality. On \texttt{hipe-2020}, Gemma4-31B achieves the strongest recall for French and German, while GPT-120B performs best for English. On \texttt{newseye}, GPT-120B is strongest for Swedish and Finnish, whereas Qwen3-A30B obtains the best recall for French and German but performs poorly on the Nordic languages. 

Overall, we notice that the retrieval performance varies substantially across languages and datasets, suggesting that multilingual coverage, training data composition, and robustness to historical spelling variation are as important as model size.

% Table~\ref{tab:retrieval_recall_base_llms} compares candidate retrieval recall across backbone LLMs without alias augmentation, isolating each model’s intrinsic parametric retrieval ability. All models are evaluated using the same simple prompting strategy (Appendix~\ref{sec:appendix}).

% Larger models generally achieve higher recall, although scaling trends are not strictly monotonic. In particular, GPT-120B is not consistently superior to smaller but more recent models such as Gemma4-31B and Qwen3.5-27B. This suggests that retrieval quality depends not only on model scale, but also on training data recency and multilingual knowledge coverage. Overall, Gemma4-31B provides the most stable cross-lingual performance, while GPT-120B remains strongest on the more challenging \texttt{newseye} setting.

\paragraph{Prompt Instruction and Alias Lookup.}
We further study the impact of prompt design, context representation, and alias augmentation. Table~\ref{tab:retrieval_recall_prompt} shows that simple prompts consistently outperform more complex instructions, while chunk-based context is more effective than summary-based context. Summary representations occasionally omit useful information, leading to lower recall. Alias lookup provides the largest improvement, consistently recovering entities missed by the LLM retrieval stage. Thus, combining simple prompts, chunk context, and alias augmentation yields the best retrieval performance.

\begin{table}[ht]
\centering
\small
\begin{tabular}{lp{.7cm}lcccc}
\toprule
\textbf{Prompt} & \textbf{Context} & \textbf{Alias} 
& \multicolumn{2}{c}{\texttt{hipe-2020}} 
& \multicolumn{2}{c}{\texttt{newseye}} \\
\cline{4-7}
 & & 
& \textbf{FR} & \textbf{DE} 
& \textbf{FR} & \textbf{DE} \\
\midrule

% Simple Prompt & Summary-oriented entity & \ding{55} 
% &  & -- & -- & -- \\
% Simple Prompt & Summary-oriented entity & \ding{51} 
% & -- & -- & -- & -- \\

% % Simple Prompt & Chunk Context & \ding{51} 
% % & 72  & 67.6 & 60.3 & 55.6 \\
Simple & Summary  & \ding{55} 
& 74.5 & 69.3 & 55.6 & 50.3\\
% Simple Prompt & Summary Context & \ding{51} 
% &   &  &  &  \\
% Complex Prompt (Simple Prompt + Meta + Entity Type)  & Chunk Context & \ding{55}  
% & 73.3 & 71.8 & 53.4 & 47.4 \\
Complex  & Chunk  & \ding{55}  
& 73.3 & 71.8 & 53.4 & 47.4 \\
Simple & Chunk  & \ding{55} 
& 75 & 72.9 & 55.8 & 50.1 \\
Simple & Chunk  & \ding{51} 
 & \textbf{76.5} & \textbf{74.4} & \textbf{57}   &  \textbf{53}\\
%Complex Prompt (Simple Prompt + Meta + Entity Type)    & Chunk Context & \ding{51}  
% & 71 & 67.1 & 51.9 & 49.4 \\

\bottomrule
\end{tabular}
\caption{Retrieval recall using different prompts, context information, and alias dictionary lookup.}
\label{tab:retrieval_recall_prompt}
\end{table}

% \subsubsection{Selection Performance}
% \begin{table*}[t]
% \centering
% \small
% \begin{tabular}{lrrrrrrr}
% \toprule
% & \multicolumn{3}{c}{\texttt{hipe-2020}} & \multicolumn{4}{c}{\texttt{newseye}} \\
% \textbf{Model} & \textbf{FR} & \textbf{DE} & \textbf{EN} & \textbf{FR} & \textbf{DE} & \textbf{SV} & \textbf{FI} \\
% \midrule
% Qwen3-14B \cite{yang2025qwen3} & 67.0 & 61.0 & 61.1 & 58.3 & 50.0 & 51.8 & 61.7 \\
% Finetuned Qwen3-14B & 67.3 & 51.8 & 65.9 & 69.1 & 62.6 & 64.0 & 60.6 \\
% \hline
% GPT-20B & 65.0 & 58.3 & 59.9 & 59.0 & 49.7 & 59.3 & 60.3 \\
% Finetuned GPT-20B \cite{agarwal2025gpt} & 72.4 & 65.6 & 73.8 & 68.7 & 59.2 & 65.0 & 62.7 \\
% \hline
% Qwen3-32B \cite{yang2025qwen3} & 67.0 & 58.5 & 63.9 & 58.7 & 51.1 & 55.2 & 60.2 \\
% Finetuned Qwen3-32B & 63.1 & 59.0 & 68.7 & 49.0 & 48.5 & 40.7 & 49.5 \\
% \bottomrule
% \end{tabular}
% \caption{Micro-F1 with different entity-selection backbones before and after finetuning.}
% \label{tab:selection_f1_llms}
% \end{table*}

\begin{table}[ht]
\centering
\small
\setlength{\tabcolsep}{4pt}
\renewcommand{\arraystretch}{0.95}
\begin{tabular}{p{4.2cm}rrr}
\toprule
& \multicolumn{3}{c}{\texttt{hipe-2020}} \\
\textbf{Model} & \textbf{FR} & \textbf{DE} & \textbf{EN} \\
\midrule
Qwen3-14B \cite{yang2025qwen3} & 67.0 & 61.0 & 61.1 \\
Finetuned MDPO Qwen3-14B & 67.3 & 51.8 & 65.9 \\
\midrule
GPT-20B \cite{agarwal2025gpt} & 65.0 & 58.3 & 59.9 \\
Finetuned MDPO GPT-20B  & 72.4 & 65.6 & 73.8 \\
\midrule
Qwen3-32B \cite{yang2025qwen3} & 67.0 & 58.5 & 63.9 \\
Finetuned MDPO Qwen3-32B & 63.1 & 59.0 & 68.7 \\
\bottomrule
\end{tabular}
\caption{Micro-F1 on hipe-2020 with different entity-selection backbones before and after fine-tuning.}
\label{tab:selection_f1_hipe}
\end{table}

\begin{table}[ht]
\centering
\small
\setlength{\tabcolsep}{3pt}
\renewcommand{\arraystretch}{0.95}
\begin{tabular}{p{4.2cm}rrrr}
\toprule
& \multicolumn{4}{c}{\texttt{newseye}} \\
\textbf{Model} & \textbf{FR} & \textbf{DE} & \textbf{SV} & \textbf{FI} \\
\midrule
Qwen3-14B \cite{yang2025qwen3} & 58.3 & 50.0 & 51.8 & 61.7 \\
Finetuned Qwen3-14B & 69.1 & 62.6 & 64.0 & 60.6 \\
\midrule
GPT-20B \cite{agarwal2025gpt} & 59.0 & 49.7 & 59.3 & 60.3 \\
Finetuned GPT-20B  & 68.7 & 59.2 & 65.0 & 62.7 \\
\midrule
Qwen3-32B \cite{yang2025qwen3} & 58.7 & 51.1 & 55.2 & 60.2 \\
Finetuned Qwen3-32B & 49.0 & 48.5 & 40.7 & 49.5 \\
\bottomrule
\end{tabular}
\caption{Micro-F1 on \texttt{newseye} with different entity-selection backbones before and after fine-tuning.}
\label{tab:selection_f1_newseye}
\end{table}

% \begin{table*}[t]
% \centering
% \small
% \begin{tabular}{lrrrrrrr}
% \toprule
% & \multicolumn{3}{c}{\texttt{hipe-2020}} & \multicolumn{4}{c}{\texttt{newseye}} \\
% \textbf{Candidate model} & \textbf{FR} & \textbf{DE} & \textbf{EN} & \textbf{FR} & \textbf{DE} & \textbf{SV} & \textbf{FI} \\
% \midrule
% Qwen3.5-9B \cite{qwen35blog} & 71.5 & 63.1 & 67.0 & 67.5 & 56.5 & 62.1 & \textbf{64.5} \\
% GPT-20B \cite{agarwal2025gpt} & 67.5 & 58.5 & 62.5 & 65.9 & 55.5 & 61.5 & 63.1 \\
% Qwen3.5-27B \cite{qwen35blog} & 68.7 & 64.1 & 72.4 & 63.6 & 57.3 & 58.7 & 59.3 \\
% Qwen3-A30B \cite{yang2025qwen3} & 62.0 & 64.2 & 72.5 & 65.0 & 58.9 & 62.5 & 60.9 \\
% Gemma4-31B \cite{gemma4_2026} & 72.1 & \textbf{65.8} & 73.0 & 65.0 & 56.3 & 59.6 & 60.2 \\
% GPT-120B \cite{agarwal2025gpt} & \textbf{72.4} & 65.6 & \textbf{73.8} & \textbf{68.7} & \textbf{59.2} & \textbf{65.0} & 62.7 \\
% \bottomrule
% \end{tabular}
% \caption{Micro-F1 for different candidate-generation models with the GPT-20B multi-DPO ranker fixed.}
% \label{tab:retrieval_f1_retrieval_llms}
% \end{table*}
\subsection{Selection Analysis}

\paragraph{Backbone LLMs.}
Tables ~\ref{tab:selection_f1_hipe} and ~\ref{tab:selection_f1_newseye} show that larger models do not necessarily yield better finetuned performance. While Qwen3-32B achieves competitive zero-shot results, its finetuned variant degrades substantially on several \texttt{newseye} languages, suggesting optimisation instability under noisy supervision. In contrast, GPT-20B consistently benefits from finetuning and achieves the best overall performance, indicating stronger compatibility with preference optimisation objectives.

\paragraph{Retrieval LLMs. }

\begin{table}[ht]
\centering
\small
\setlength{\tabcolsep}{4pt}
\renewcommand{\arraystretch}{0.95}
\begin{tabular}{lrrr}
\toprule
& \multicolumn{3}{c}{\texttt{hipe-2020}} \\
\textbf{Candidate model} & \textbf{FR} & \textbf{DE} & \textbf{EN} \\
\midrule
Qwen3.5-9B \cite{qwen35blog} & 71.5 & 63.1 & 67.0 \\
Qwen3.5-27B \cite{qwen35blog} & 68.7 & 64.1 & 72.4 \\
Qwen3-A30B \cite{yang2025qwen3} & 62.0 & 64.2 & 72.5 \\
GPT-20B \cite{agarwal2025gpt} & 67.5 & 58.5 & 62.5 \\
GPT-120B \cite{agarwal2025gpt} & \textbf{72.4} & 65.6 & \textbf{73.8} \\
Gemma4-31B \cite{gemma4_2026} & 72.1 & \textbf{65.8} & 73.0 \\
\bottomrule
\end{tabular}
\caption{Micro-F1 on \texttt{hipe-2020} with the same multi-negative  finetuned DPO model}
\label{tab:retrieval_f1_hipe}
\end{table}

\begin{table}[ht]
\centering
\small
\setlength{\tabcolsep}{3pt}
\renewcommand{\arraystretch}{0.95}
\begin{tabular}{lrrrr}
\toprule
& \multicolumn{4}{c}{\texttt{newseye}} \\
\textbf{Candidate model} & \textbf{FR} & \textbf{DE} & \textbf{SV} & \textbf{FI} \\
\midrule
Qwen3.5-9B \cite{qwen35blog} & 67.5 & 56.5 & 62.1 & \textbf{64.5} \\
Qwen3.5-27B \cite{qwen35blog} & 63.6 & 57.3 & 58.7 & 59.3 \\
Qwen3-A30B \cite{yang2025qwen3} & 65.0 & 58.9 & 62.5 & 60.9 \\
GPT-20B \cite{agarwal2025gpt} & 65.9 & 55.5 & 61.5 & 63.1 \\
GPT-120B \cite{agarwal2025gpt} & \textbf{68.7} & \textbf{59.2} & \textbf{65.0} & 62.7 \\
Gemma4-31B \cite{gemma4_2026} & 65.0 & 56.3 & 59.6 & 60.2 \\
\bottomrule
\end{tabular}
\caption{Micro-F1 on \texttt{newseye} with the same multi-negative  DPO model}
\label{tab:retrieval_f1_newseye}
\end{table}

Tables ~\ref{tab:retrieval_f1_hipe} and ~\ref{tab:retrieval_f1_newseye}  evaluate  the impact of retrieval quality while keeping the same GPT-20B multi-DPO ranker fixed. Stronger retrieval models generally improve end-to-end performance, with GPT-120B achieving the best results overall. However, the gains remain moderate, suggesting that the ranker can partially compensate for weaker candidate generation. Mid-sized models such as Gemma4-31B and Qwen3.5-27B remain competitive across languages.

\subsection{Error Analysis}
\begin{figure}
    \centering
    \includegraphics[width=0.9\linewidth]{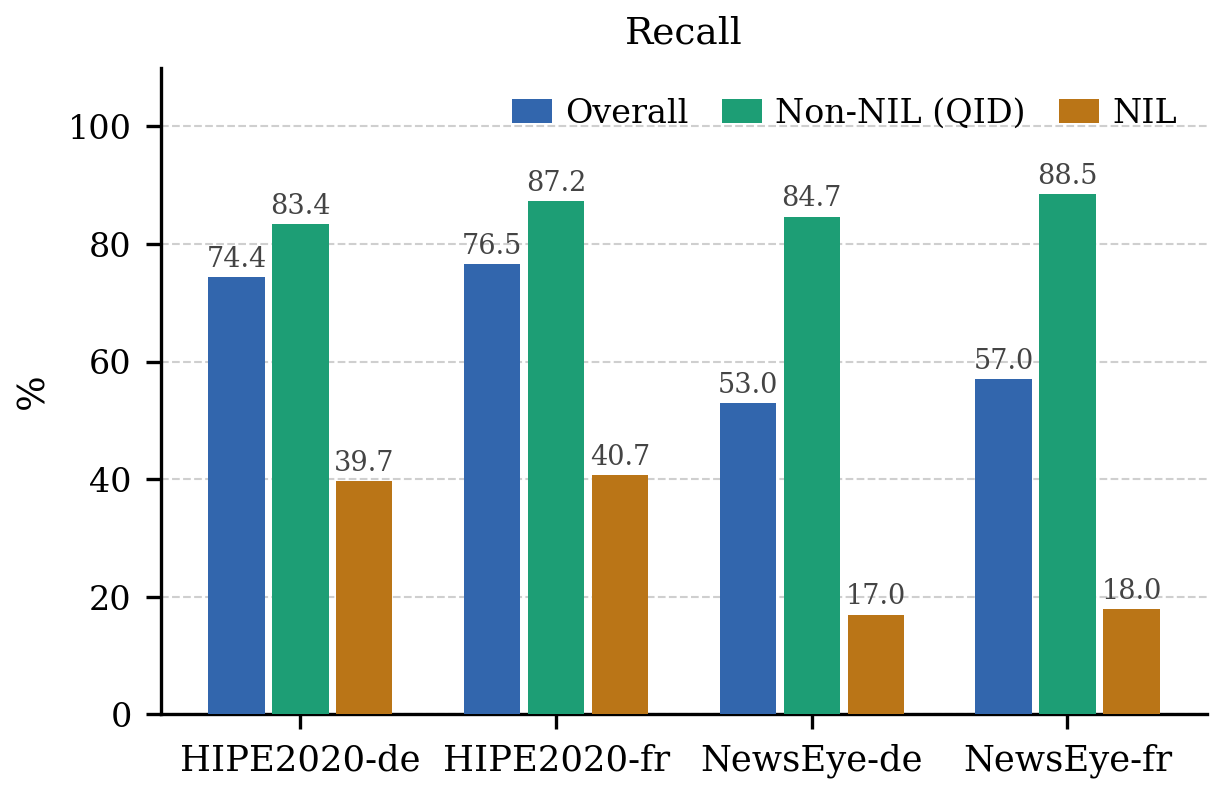}
    \caption{Retrieval recall (\%) across datasets.}
    \label{fig:retrieval_recall}
\end{figure}

\begin{figure}[t]
    \centering
    \begin{subfigure}[t]{0.9\linewidth}
        \centering
        \includegraphics[width=\linewidth]{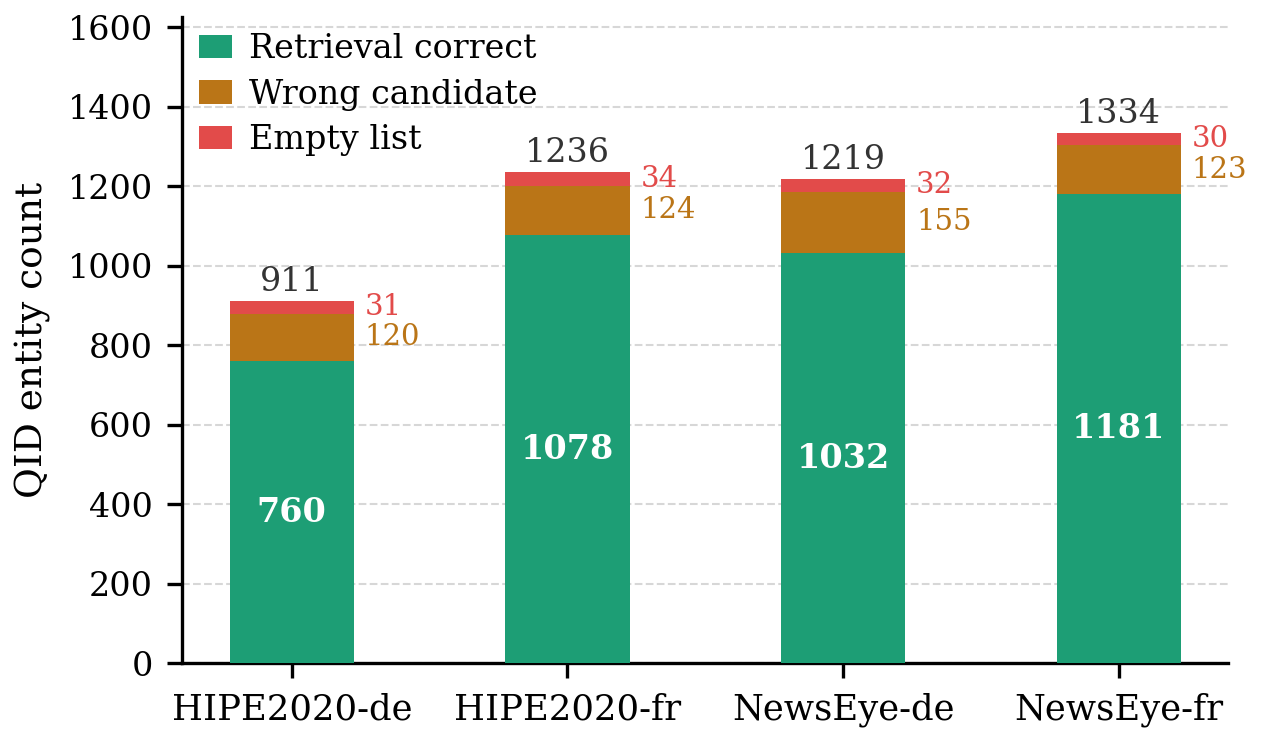}
        \caption{Simple prompt}
    \end{subfigure}
    \vspace{0.5em}
    \begin{subfigure}[t]{1.08\linewidth}
        \centering
        \includegraphics[width=\linewidth]{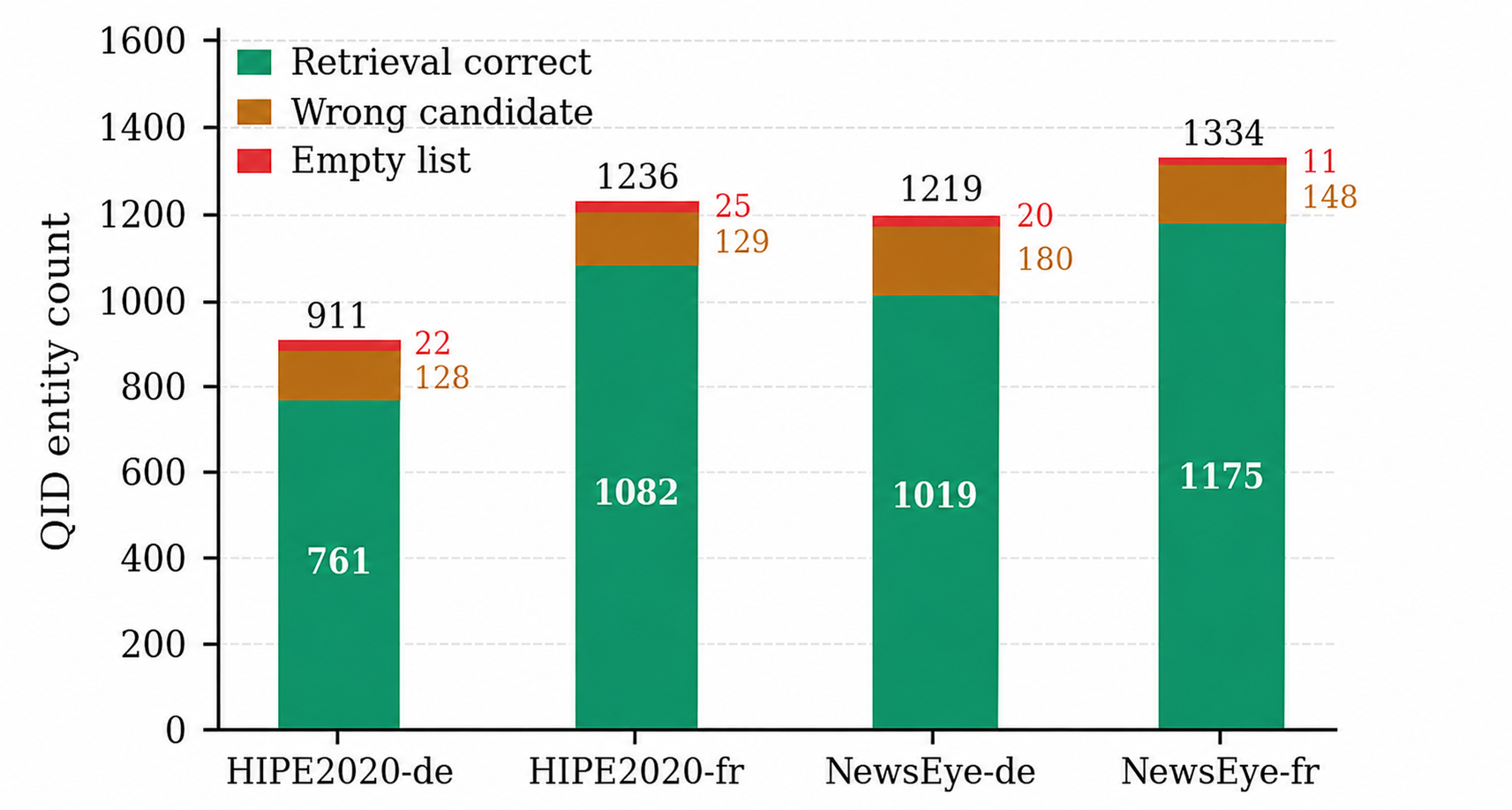}
        \caption{Complex prompt}
    \end{subfigure}
    \caption{Retrieval error breakdown for simple and complex prompts.}
    \label{fig:retrieval_errors}
\end{figure}

\paragraph{Retrieval errors.}
This section provides a detailed analysis of candidate generation and its interaction with downstream entity selection. Across the four subsets shown in Figure~\ref{fig:retrieval_recall},  non-NIL recall ranges from 83.4\% on \texttt{hipe-2020}-DE to 88.5\% on \texttt{newseye}-FR. In contrast, NIL recall ranges only from 17.0\% on \texttt{newseye}-DE to 40.7\% on \texttt{hipe-2020}-FR because the system often retrieves a plausible entity instead of abstaining. NIL handling therefore remains the main retrieval weakness. 
\label{app:retrieval-analysis}
Figure~\ref{fig:retrieval_errors} compares simple and complex prompts: simple prompts generally recover more correct candidates, whereas complex prompts may reduce empty outputs at the cost of generating additional incorrect candidates. %These findings motivate analysing retrieval and selection as separate sources of error.

\paragraph{Fine-Grained Errors.} We heuristically categorise errors across all seven subsets and report the absolute accuracy gain of the finetuned selector over the base model in Table~\ref{tab:error_type_gain}. The largest gains occur for gold-NIL/knowledge-base coverage (+13.7\%) and semantic ambiguity (+8.1\%), followed by OCR noise. These are relative improvements, not solved cases: low absolute NIL retrieval recall and false NIL predictions on valid QID mentions remain important limitations. Appendix~\ref{sec:appendix_error_analysis} provides additional examples involving historical spelling, stub entities, fine-grained geography, and evolving gold-NIL annotations.
\begin{table}[ht]
\centering
\small
\setlength{\tabcolsep}{2.0pt}
\begin{tabular}{lrrrr}
\toprule
\textbf{Error type} & $N$ & \textbf{Base} & \textbf{Finetuned} & \textbf{Gain (pp)} \\
\midrule
NIL / KB & 3636 & 48.4 & 62.1 & +13.7 \\
Semantic & 801 & 59.4 & 67.5 & +8.1 \\
OCR noise & 79 & 53.2 & 60.8 & +7.6 \\
Other & 2993 & 68.2 & 71.8 & +3.6 \\
Spelling / orthog. & 1194 & 63.7 & 64.9 & +1.2 \\
Historical naming & 717 & 56.5 & 57.5 & +1.0 \\
\bottomrule
\end{tabular}
\caption{Accuracy by heuristic error type across the seven evaluation subsets. ``Other'' contains cases not matched by the surface-form heuristics.}
\label{tab:error_type_gain}
\end{table}

\section{Conclusions}\label{sec:conclusion}
In this work, we introduced a retrieval-augmented framework for historical named entity linking that combines LLM-based candidate generation with preference-optimised entity selection. We showed that while modern LLMs provide strong retrieval capabilities, effective candidate ranking remains the primary challenge in multilingual and OCR-degraded historical documents. To address this, we proposed a multi-negative DPO objective that improves the model’s ability to distinguish correct entities from noisy candidates. Experiments on \texttt{hipe-2020} and \texttt{newseye} demonstrate consistent improvements over both supervised and unsupervised baselines across multiple languages. Our ablation studies further show that retrieval quality, context design, and preference optimization all contribute to final performance, while moderate context windows and simple prompting strategies are generally the most effective.

% \clearpage
\section*{Limitations}

Although our LLM-guided retrieval framework improves candidate recall in noisy historical corpora, several limitations remain.

First, the approach still struggles with NIL entity prediction. The LLM tends to generate semantically plausible entities even when no correct entity exists in the knowledge base, leading to increased false positives. This behavior is particularly challenging in historical documents containing OCR errors, spelling variations, and incomplete contextual information.

Second, the overall performance remains highly dependent on the retrieval stage. If the correct entity is not retrieved or validated, the downstream selection model cannot recover it. While LLM-guided query expansion improves recall, it may also introduce noisy or hallucinated candidates.

Third, our objective is a decomposable pairwise extension of single-negative DPO. We did not evaluate alternative listwise or contrastive objectives with shared candidate-set normalisation, so our results do not establish superiority over those formulations.

Finally, candidate generation is the dominant inference cost in the measured deployable configuration, and the reported latency is not directly comparable with published systems running on different hardware and software stacks.

\section*{Acknowledgments}
This work has been co-funded by the European Union HORIZON-WIDERA-2023-TALENTS-01-01 grant 101186647 — AI4DH. Views and opinions expressed are however those of the author(s) only and do not necessarily reflect those of the European Union. Neither the European Union nor the granting authority can be held responsible for them. 

 This work has also been supported by the TERMITRAD (2020-2019-8510010) and ACTUADATA (2022-2021-17014610) projects funded by the Nouvelle-Aquitaine Region (France), and it has benefited from the computing resources of the L3i laboratory, operated and hosted by the University of La Rochelle, and funded by the French government and the Nouvelle-Aquitaine Region.
%\bibliographystyle{acl_natbib}
\bibliography{custom}

\appendix
\section{Training and Inference Details}\label{app:training-details}
We subclass the TRL DPO trainer\footnote{\url{https://huggingface.co/docs/trl/dpo_trainer}} and use a custom collator so that every EL instance retains one prompt, one chosen entity, and all rejected candidates from the same retrieval set. %The collator pads rejected responses to $[B,N,L]$ and records the number of valid negatives per instance. 
The trainer computes length-normalised policy and reference log-likelihoods, obtains the chosen score once, compares it independently with every valid rejected score, and masks padded entries before averaging. The frozen reference is the pretrained backbone with LoRA adapters disabled. Thus, the code implements Eq.~\ref{eq:mdpo} rather than flattening examples into unrelated one-to-one records.

LoRA is applied to the attention and feed-forward projection layers %(\texttt{q\_proj}, \texttt{k\_proj}, \texttt{v\_proj}, \texttt{o\_proj}, \texttt{gate\_proj}, \texttt{up\_proj}, and \texttt{down\_proj}) 
with rank $r=8$ and $\alpha=16$. 

Training is performed for 3 epochs using AdamW 8-bit optimisation with a learning rate of $5 \times 10^{-6}$, linear learning-rate scheduling, and a warmup ratio of 0.1. We use a per-device batch size of 8 with gradient accumulation over 8 steps. Models are trained in BF16 precision with gradient checkpointing enabled. The model context window is 2048 tokens; prompts are capped at 1024 tokens and answers at 128 tokens. 
%The resulting risk is information loss through truncation, not context overflow. 
The LLM-generated branch is capped at five candidates; larger alias-augmented pools require candidate pruning or description truncation.

For inference, the merged LoRA model is quantised to 4-bit MXFP4 format and served using vLLM\footnote{\url{https://vllm.ai}} for efficient batched decoding and memory-efficient inference. On one RTX A6000, a deployable GPT-20B generator/selector configuration requires 2.35--2.63~s for candidate generation and 0.35--0.50~s for selection (2.70--3.13~s combined). Candidate generation accounts for approximately 82--89\% of latency. The strongest configuration in Table~\ref{tab:sota_hipe_newseye} instead uses a GPT-120B generator; because published systems use different hardware and implementations, we do not make a direct runtime claim against them.

We follow the official dataset splits for all experiments. For \texttt{hipe-2020}, we train on the French and German training sets and use the corresponding development sets for validation. The English subset is excluded from training and evaluated only in a zero-shot cross-domain setting to assess generalisation on unseen historical data. For \texttt{newseye}, we use the French, German, Swedish, and Finnish subsets following the same train/development split protocol. 

For multi-negative preference construction, we use GPT-20B \cite{agarwal2025gpt} as the candidate generation model, served through the vLLM inference engine for efficient large-scale decoding. For the standard DPO baseline, preference pairs are constructed using the gold entity as the positive sample and a randomly selected entity from the negative candidate pool as the negative sample. 

Instances without a valid positive/negative comparison are excluded from DPO training. At inference, a single candidate is not automatically accepted because the generative selector may still emit NIL or an invalid response. In a controlled test, the selector chose the sole gold candidate in 44/50 cases (88.0\%); for 50 empty-candidate gold-NIL cases, it returned NIL in every case.

Although GPT-20B \cite{agarwal2025gpt} generates high-quality multi-negative comparison pairs, its SFT convergence was unstable. We therefore use Qwen3-14B \cite{yang2025qwen3} for the SFT baseline.

\section{Selection}
\subsection{Error Analysis Across Retrieval and Selection}
\begin{figure}[t]
    \centering
    \includegraphics[width=\linewidth]{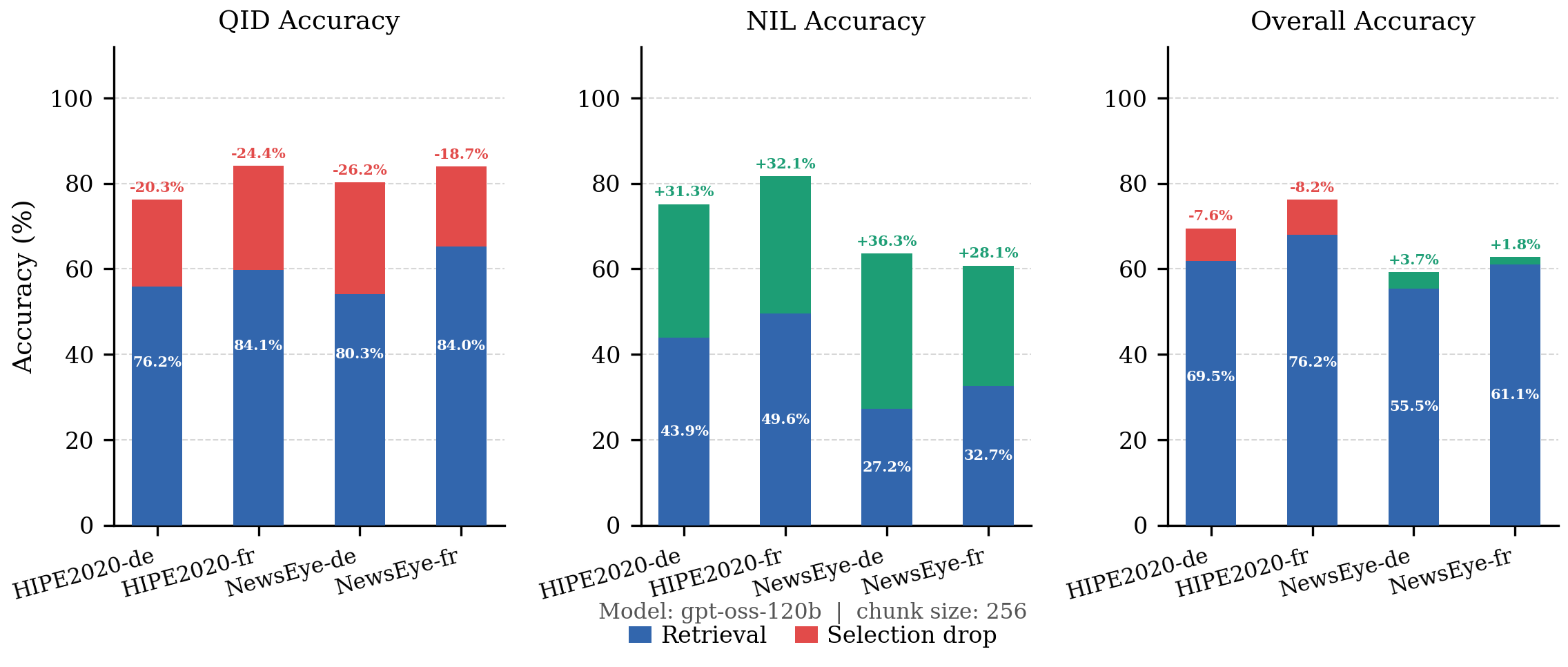}
    \caption{Entity-level accuracy before and after selection.}
    \label{fig:accuracy_selection_llms}
\end{figure}

Although the primary evaluation metric is micro-F1, we report entity-level accuracy to better analyse the transition from retrieval to selection. Figure~\ref{fig:accuracy_selection_llms} compares performance before and after the selection stage. We observe that the selection model consistently improves NIL prediction accuracy across all datasets, with substantial gains (up to +36\%), indicating its effectiveness in identifying non-linkable mentions. 

However, this improvement comes at the cost of reduced QID accuracy, where selection introduces a noticeable drop (up to -26\%). As a result, overall accuracy shows only modest changes, with slight improvements in some cases and small degradations in others. These results highlight a key trade-off: while the selection model enhances robustness in handling NIL cases, it can also over-filter valid candidates, suggesting that balancing precision and recall in the selection stage remains critical.
\subsection{Context length.}
\begin{table}[ht]
\centering
\small
\begin{tabular}{lcccc}
\hline
\textbf{Context length} 
& \multicolumn{2}{c}{\texttt{hipe-2020}} 
& \multicolumn{2}{c}{\texttt{newseye}} \\
\cline{2-5}
& \textbf{FR} & \textbf{DE} & \textbf{FR} & \textbf{DE} \\
\hline
128 & 64.4 & 59.1 & 64.1 & 53.9 \\
256 & \textbf{72.4} & \textbf{65.6} & \textbf{68.7} & \textbf{59.2} \\
512 & 69.8 & 64.7 & 67.9 & 56.6 \\
\hline
\end{tabular}
\caption{F1 score with different context lengths for backbone LLMs.}
\label{tab:selection_f1_llms_contextlength}
\end{table}

Table~\ref{tab:selection_f1_llms_contextlength} shows that a moderate context window (256 tokens) yields the best overall performance. Shorter contexts lack sufficient disambiguating information, while longer contexts (512 tokens) often introduce additional noise and slightly degrade performance. This suggests a trade-off between contextual coverage and irrelevant information in retrieval-augmented selection.
\section{Instruction Prompts}
\label{sec:appendix}

\subsection{Simple Retrieval Prompt}
\label{app:retrieval_prompt}

The following prompt in Tab ~\ref{tab:retrieval_prompt} was used to generate candidate entity names for Wikidata entity linking.

\noindent
\begin{table}[t]
\centering
\caption{Complete retrieval prompt used for candidate generation.}
\label{tab:retrieval_prompt}

\begin{tabularx}{\columnwidth}{|>{\raggedright\arraybackslash\ttfamily\scriptsize}X|}
\hline

RETRIEVAL\_PROMPT = """

You are an expert in Wikidata entity linking and historical linguistics.
Your goal is to generate a concise list of potential candidate entity names
based on the given context.

Attention: \\
- The entity mention may include typos, punctuation, extra spaces, or unusual capitalisation. \\
- The surrounding context text is useful to help identify the entity's historical or cultural background. \\

Your task: \\
1. Normalise the entity string,  remove unnecessary punctuation, spacing, or symbols. \\
2. Use contextual clues, such as time period, nationality, and occupation, to infer more precise or alternate forms. \\
3. Generate up to **5 likely candidate names** that could correspond to valid Wikidata items. \\
4. Each candidate must be a clean, human-readable string and validated through the wikidata\_checker function.

Rules:\\
- Do not use markdown, extra commentary, or natural language. \\
- If no plausible candidates are found, return the original entity.
\\
Mode: No Reasoning

"""\\

\hline
\end{tabularx}
\end{table}

\subsection{Complex Retrieval Prompt}
\label{app:retrieval_complex_prompt_2}

The following prompt in Tab ~\ref{tab:retrieval_complex_prompt} was used for handling complex historical multilingual text retrieval and OCR noise:

\begin{table*}[t]
\centering
\caption{Complex retrieval prompt used for Wikidata candidate generation.}
\label{tab:retrieval_complex_prompt}
\setlength{\tabcolsep}{6pt}
\renewcommand{\arraystretch}{1.05}

\begin{tabularx}{\textwidth}
{|>{\raggedright\arraybackslash\ttfamily\footnotesize}X|}
\hline

COMPLEX\_RETRIEVAL\_PROMPT = f"""\\[2pt]

You are an expert in Wikidata entity linking for historical multilingual texts.\\
The entity mentions come from OCR-scanned historical newspapers
(French, German) published between 1797 and 1920, from Swiss,
Luxembourgish, and Austrian collections.\\[2pt]

You will be given the entity mention, its type
(PERS/LOC/ORG/PROD/TIME), and context.\\
Generate up to \{MAX\_CANDIDATES\} candidate name strings to search on Wikidata.\\[2pt]

\#\# By entity type:\\
- PERS \textrightarrow{} search the surname first, then the full name;
drop honorifics and titles\\
- LOC \textrightarrow{} strip generic geographic words; search the proper name\\
- ORG \textrightarrow{} search the canonical short name of the organisation\\
- PROD \textrightarrow{} search the publication or doctrine title\\
- TIME \textrightarrow{} return empty (dates are not linkable to Wikidata)\\[2pt]

\#\# Common OCR noise in these texts:\\
- Broken hyphens, extra spaces, or stray characters from imperfect scanning\\
- Abbreviated honorifics and titles --- expand or drop them\\
- Old or non-standard spellings --- try the modern standard form\\
- Multi-token mentions where only part is the actual entity name\\[2pt]

\#\# Steps:\\
1. Clean OCR noise from the mention\\
2. Use the entity type to identify the core searchable part\\
3. Generate variants: clean form, sub-parts, expanded abbreviations,
and alternate spellings\\
4. Order candidates: most distinctive first, raw cleaned mention last\\[2pt]

Rules:\\
- Validate each candidate through the wikidata\_checker function\\
- Return clean human-readable strings only --- no QIDs and no markdown\\
- Use context clues such as dates, nationality, and profession\\
- If no candidates are found, return the cleaned raw mention\\[2pt]

Mode: No Reasoning\\[2pt]

"""\\

\hline
\end{tabularx}
\end{table*}

\subsection{Selection Prompt}
\label{app:selection_prompt}

The following prompt in Tab ~\ref{tab:selection_prompt} was used for Named Entity Linking, given the context and candidate strings:

\noindent
\begin{table}[t]
\centering
\caption{Selection prompt used for entity linking.}
\label{tab:selection_prompt}

\setlength{\tabcolsep}{5pt}
\renewcommand{\arraystretch}{1.05}

\begin{tabularx}{\columnwidth}
{|>{\raggedright\arraybackslash\ttfamily\scriptsize}X|}
\hline

SELECTION\_PROMPT = """\\

You are an expert Named Entity Linking (NEL) specialist. You are given an entity, its context, and a list of candidate strings.

Return:\\
\textless entity\textgreater: \textless QID\textgreater\\

Rules:\\
- If no candidate is suitable, return ``NIL''.
- Only use provided candidates; do not invent new ones.

"""\\

\hline
\end{tabularx}
\end{table}

% EMNLP 2026 Appendix - Named Entity Linking Error Analysis

% EMNLP 2026 Appendix - Named Entity Linking Error Analysis
\section{Detailed Error Analysis}
\label{sec:appendix_error_analysis}

To better understand the behavior of our approach, we conduct a qualitative error analysis across the seven historical entity linking datasets from \texttt{hipe-2020} and \texttt{newseye}.

\subsection{Success Cases and Comparison with Baseline LLM}

Finetuning substantially mitigates several limitations observed in zero-shot LLM-based entity linking. In particular, baseline LLMs frequently struggle with historical spelling variation, ambiguity introduced by duplicate or incomplete Wikidata entries, and fine-grained geographical disambiguation. For example, the baseline model often links \textit{Bartenstein} to a deprecated or stub Wikidata entry, or resolves \textit{Philadelphia} to the naval ship rather than the city. Similar errors occur for geographically ambiguous mentions such as \textit{Berne} and \textit{New-York}, where the model confuses cities with larger administrative regions.

As illustrated in Table~\ref{tab:success_cases}, the finetuned model learns stronger context-sensitive linking preferences and better captures historical newspaper conventions, enabling more accurate selection of the gold-standard Wikidata entities.

\begin{table*}[t]
\centering
\small
\setlength{\tabcolsep}{4pt}
\renewcommand{\arraystretch}{1.12}

\newcommand{\qid}[1]{\texttt{#1}}

\begin{tabularx}{\textwidth}{
  l
  >{\raggedright\arraybackslash}X
  >{\centering\arraybackslash}p{2.1cm}
  >{\raggedright\arraybackslash}p{2.9cm}
  >{\raggedright\arraybackslash}p{2.9cm}
}
\toprule
\textbf{Dataset} &
\textbf{Context window} &
\makecell[c]{\textbf{Gold}\\\textbf{entity}} &
\makecell[l]{\textbf{Baseline LLM}\\\textbf{without FT}} &
\makecell[l]{\textbf{Finetuned}\\\textbf{model}} \\
\midrule

\texttt{HIPE2020}-DE & ... Am 22 . Merz sah man auf der großen Parade zu \textbf{Petersburg} 4 . eroberte schwedische Fahnen... & Saint Petersburg (\texttt{Q656}) & \texttt{NIL} & Saint Petersburg (\texttt{Q656}) \\
\texttt{HIPE2020}-FR & ... lieutenant d ' Avoyer de la Préfecture de Berne , à \textbf{Berne} le 30 Avril 1828... & Bern (\texttt{Q70}) & Canton of Berne (\texttt{Q11911}) & Bern (\texttt{Q70}) \\
\texttt{HIPE2020}-FR& ... Les amateurs pourront se trouver rassemblés à l ' auberge du Cerf aux \textbf{Ponts} . & Les Ponts-de-Martel (\texttt{Q68617}) & Les Ponts-de-Cé (\texttt{Q752975}) & Les Ponts-de-Martel (\texttt{Q68617}) \\
\texttt{HIPE2020}-FR & ... soignée du beau Traire des causes civiles de la Principauté de \textbf{Neuchâtel} , par Al . le châtelain Alonvert ... & Neuchâtel (\texttt{Q69345}) & Principality of Neuchâtel (\texttt{Q3137802}) & Neuchâtel (\texttt{Q69345}) \\
\texttt{HIPE2020}-EN & ... ship Caledonia , arrived at \textbf{Philadelphia} on Monday , from Cadiz , states , that the French army ... & Philadelphia (\texttt{Q1345}) & USS Philadelphia (\texttt{Q2288745}) & Philadelphia (\texttt{Q1345}) \\
\texttt{HIPE2020}-EN & ... respectable meeting in \textbf{New - York} , against the violation of the treaties with the... & New York City (\texttt{Q60}) & New York State (\texttt{Q1384}) & New York City (\texttt{Q60}) \\
\texttt{HIPE2020}-EN & ... while that proposed by \textbf{Gen . Harrison} , with the present number ot our militia , would cost... & William H. Harrison (\texttt{Q11869}) & W. H. Harrison (politician) (\texttt{Q8012023}) & William H. Harrison (\texttt{Q11869}) \\
\texttt{newseye}-DE & ... Plener wünschen , wogegen bei der Wahl im Jahre 1891 Herr \textbf{v . Plener} einen Gegenkandidaten hatte ... & Ernst von Plener (\texttt{Q324918}) & von Plener (family) (\texttt{Q23866513}) & Ernst von Plener (\texttt{Q324918}) \\
\texttt{newseye}-DE& ... des Satanismus sadistische Befriedigung sucht . Der Satanismus wird Modesache . Unter \textbf{Ludwig XIV .} wird die schwarze Messe populär... & Louis XIV of France (\texttt{Q7742}) & \texttt{NIL} & Louis XIV of France (\texttt{Q7742}) \\ 
\texttt{newseye}-FR  & ... Brest , 14 janvier . — Le \textbf{conseil municipal} de Camaret vient de démissionner , parce qu ' il... & municipal council (\texttt{Q701632}) & city council (\texttt{Q3154693}) & municipal council (\texttt{Q701632}) \\

\bottomrule
\end{tabularx}

\caption{Examples of named entity linking where the baseline LLM without finetuning fails (by predicting incorrect/stub QIDs or returning \texttt{NIL}), but our proposed finetuned model correctly links to the gold standard Wikidata QID.}
\label{tab:success_cases}
\end{table*}

\subsection{Wikidata Annotation Discrepancies (Gold NIL vs.\ Predicted QID)}

A recurring challenge in historical entity linking is the evolving nature of the underlying knowledge base. Benchmark datasets such as \texttt{hipe-2020} and \texttt{newseye} rely on static annotations, where many mentions are labeled as \texttt{NIL} because no suitable Wikidata entity was available or verified at annotation time.

However, Wikidata has expanded considerably since the creation of these datasets. Combined with dense alias retrieval, our system is often able to retrieve plausible and contextually appropriate Wikidata entities for mentions annotated as \texttt{NIL}. Table~\ref{tab:discrepancies} presents representative examples where the predicted entity appears historically consistent despite disagreement with the original annotation.

\begin{table*}[t]
\centering
\small
\setlength{\tabcolsep}{4pt}
\renewcommand{\arraystretch}{1.12}

\newcommand{\qid}[1]{\texttt{#1}}

\begin{tabularx}{\textwidth}{
  l
  >{\raggedright\arraybackslash}X
  >{\centering\arraybackslash}p{1.8cm}
  >{\raggedright\arraybackslash}p{4.0cm}
}
\toprule
\textbf{Dataset} &
\textbf{Context window} &
\textbf{Gold label} &
\makecell[l]{\textbf{Model prediction}\\\textbf{(Wikidata entity)}} \\
\midrule

\texttt{HIPE2020-FR} & ... curateur , ont été admis par arrêt du \textbf{Conseil d ' Etat} du 5 décembre 1797 , à solliciter une... & \texttt{NIL} & Conseil d'État (\texttt{Q769657}) \\
\texttt{HIPE2020-FR} & ... tenoit feu Mlle . Eckard à la \textbf{rue des Moulins} , avisent le public qu ' elles le tiendront... & \texttt{NIL} & rue des Moulins (\texttt{Q3451926}) \\
\texttt{HIPE2020-EN} & ... moft obedient , and very humble fervants , \textbf{RICHARD HENRY LEE} WILLIAM GRAYSON . PRESIDENT SULLIVAN... & \texttt{NIL} & Richard Henry Lee (\texttt{Q725907}) \\
\texttt{NewsEye-DE}  & ... Kramarsch hat heute erzählt , daß Fräulein \textbf{Marie Pospischil} , eine Schauspielerin des tschechischen Theaters in Prag... & \texttt{NIL} & Marie Pospíšilová (\texttt{Q12035450}) \\
\texttt{NewsEye-DE}  & ... Klausen . Historischer Roman von \textbf{Johann von Wildenradt} . Vor dem Rathhause stieg der siegreiche... & \texttt{NIL} & Johann von Wildenrath (\texttt{Q55681870}) \\
\texttt{NewsEye-FR}  & ... Steeg , gouverneur de l ' Algérie , et \textbf{M . Rault} , président de la commission du gouvernement de la Sarre... & \texttt{NIL} & Michel Rault (\texttt{Q65597615}) \\
\texttt{NewsEye-FI}  & ... kehotettu käymään Englannissa , Saksassa ja Venäjällä puhumassa \textbf{San Franciscon} näyttelyn puolesta . & \texttt{NIL} & San Francisco (\texttt{Q62}) \\
\bottomrule
\end{tabularx}
\caption{Examples of Wikidata annotation discrepancies: mentions labeled as \texttt{NIL} in the official gold standard, which our proposed finetuned model successfully links to the active Wikidata QID.}
\label{tab:discrepancies}
\end{table*}

\subsection{Key Qualitative Insights}

\begin{itemize}

\item \textbf{Historical Orthographic Ambiguity:}
In \texttt{hipe-2020-DE}, the mention \textit{Bartenstein} refers to the historical East Prussian city now known as \textit{Bartoszyce}. The baseline model is misled by duplicate or inactive Wikidata entries and predicts an incorrect QID (\texttt{Q6467766}), whereas the finetuned model correctly resolves the mention to \textit{Bartoszyce} (\texttt{Q809585}).

\item \textbf{Fine-Grained Contextual Disambiguation:}
Baseline LLMs frequently prefer broader or more globally prominent entities. For instance, \textit{Berne} is incorrectly linked to the \textit{Canton of Berne} (\texttt{Q11911}) instead of the city (\texttt{Q70}), while \textit{New-York} is resolved to \textit{New York State} (\texttt{Q1384}) rather than \textit{New York City} (\texttt{Q60}). The finetuned model more reliably distinguishes municipalities from larger administrative regions using local contextual cues.

\item \textbf{Resolution of Gold NIL Mentions:}
Several mentions annotated as \texttt{NIL} can be linked to plausible Wikidata entities using updated knowledge base information. For example, in \texttt{NewsEye-DE}, the mention \textit{Marie Pospischil} is linked by our model to \textit{Marie Pospíšilová} (\texttt{Q12035450}), a Czech actress historically active in Prague theater. Similarly, in \texttt{NewsEye-FR}, the model links \textit{M. Rault} to the French politician (\texttt{Q65597615}). These examples suggest that benchmark annotations may underestimate real-world linking performance in historically evolving knowledge bases.

\end{itemize}

\end{document}